\documentclass{article}

\usepackage{PRIMEarxiv}

\usepackage[utf8]{inputenc} 
\usepackage[T1]{fontenc}    
\usepackage{hyperref}       
\usepackage{url}            
\usepackage{booktabs}       
\usepackage{amsfonts}       
\usepackage{nicefrac}       
\usepackage{microtype}      
\usepackage{lipsum}
\usepackage{fancyhdr}       
\usepackage{graphicx}       
\graphicspath{{media/}}     
\usepackage{pgfplots}

\title{Comparing a composite model versus chained models to locate a nearest visual object
}

\author{
  Antoine Le Borgne \\
  Orange Innovation \\
  Lannion\\
  \texttt{\ antoine.leborgne@orange.com} \\
   \And
  Xavier Marjou \\
  Orange Innovation \\
  Lannion\\
  \texttt{\ xavier.marjou@orange.com} \\
   \And
  Fanny Parzysz \\
  Orange Innovation \\
  Lannion\\
  \texttt{\ fanny.parzysz@orange.com} \\
   \And
  Tayeb Lemlouma \\
  IRISA \\
  Lannion\\
  \texttt{\ tayeb.lemlouma@irisa.fr} \\
}

\begin{document}
\maketitle

\begin{abstract}
    Extracting information from geographic images and text is crucial for autonomous vehicles to determine in advance the best cell stations to connect to along their future path. Multiple artificial neural network models can address this challenge; however, there is no definitive guidance on the selection of an appropriate model for such use cases.
    Therefore, we experimented two architectures to solve such a task: a first architecture with chained models where each model in the chain addresses a sub-task of the task; and a second architecture with a single model that addresses the whole task. 
    Our results showed that these two architectures achieved the same level performance with a root mean square error (RMSE) of 0.055 and 0.056; The findings further revealed that when the task can be decomposed into sub-tasks, the chain architecture exhibits a twelve-fold increase in training speed compared to the composite model. Nevertheless, the composite model significantly alleviates the burden of data labeling.
    
\end{abstract}

\section{Introduction}

Extracting information from geographic images and text is crucial to predict connectivity for autonomous vehicles. Autonomous vehicles require excellent cellular connectivity to interact with other connected vehicle, cars analyze images to locate themselves in their environment or even unmanned aerial vehicles (UAVs) for use-case like surveillance of asset over long geographical distances. Consequently, most operators want to assess the radio reception of their autonomous vehicle along their planned trajectories. Various connectivity-related features can be pertinent in predicting the cellular connectivity of a given area: the number of base stations, their geographic position and azimuth of the antenna, the topology of the area, the characteristics of the area (rural, urban) or the population density. Typically, these inputs come from tabular data and map images, thus requiring multi-modal models.
 
With recent advances in machine learning to perform multi-modal tasks, recent studies show that in relatively near future, Large Language models should be the most suitable model for multi-modal tasks \cite{zhang2023llamaadapter}. Large Language Models (LLMs) are neural networks with billions of parameters that are trained on vast amounts of unlabelled text using self-supervised learning. LLMs have emerged as a turning point in natural language processing research, shifting the focus away from specialized supervised models for specific tasks.  Recently, GPT4.0, a generative pre-trained Transformer (GPT) based model \cite{radford2018improving}, was announced to be a "large multi-modal model (accepting image and text inputs, emitting text outputs)"\cite{openai2023gpt4}. Although image support is not yet officially open to users, in its current state, GPT4.0 seems capable of roughly recognizing image content, but not capable of specific tasks such as measuring distance, as shown by an example in appendix. To be fair with GPT4.0, additional testing would be needed given that we only experimented it in one-shot or few-shot learning, not a full learning as with our methods.

Another LLM that has reached the state of the art is LLaMA presented by \cite{touvron2023llama}. And in recent work, \cite{zhang2023llamaadapter} presents an adapter that can be tuned in a few hours to enable LLaMA to support multi-modal input as images, it may be interesting for the future to look at this work.

However, in our work we focused on extracting information only from images. A first alternative is to divide this task into multiple sub-task and chain a sequence of models, one model dedicated to each sub-task.  But for each of the sub-tasks, this requires significant labeling work that we want to avoid. A second alternative is to use a single model that we call composite model, to entirely solve the task. To our knowledge, no study has compared whether it was better to use this composite model or chained models composed of several models specific to each sub-task.

This article focuses on the comparison of two models that aim to predict the (x, y) coordinates of the base station closest to a drone based on an input image. This prediction task involves two essential sub-tasks: information extraction from the image and subsequent analysis to determine the (x, y) coordinates. Deep learning models such as Convolutional Neural Networks (CNNs) and Transformers have demonstrated their efficacy in these areas, making them suitable choices for accomplishing these tasks effectively.

Convolutional Neural Networks (CNNs) have become increasingly popular for regression tasks in recent years \cite{bekkar2021air} \cite{oehmcke2022deep} \cite{apostolopoulos2020covid}. Their ability to learn complex feature representations from image data has made them a natural choice for tasks such as image processing. One common approach is to use pre-trained CNNs as feature extractors and feed the extracted features into a linear regression model. This has been successful in applications such as predicting air quality from satellite images \cite{bekkar2021air}, estimating forest biomass from UAV-based LiDAR data \cite{oehmcke2022deep}, or detecting COVID-19 from x-ray scans \cite{apostolopoulos2020covid}.
Another approach is to train CNNs end-to-end for regression tasks, which allows the model to learn both the feature representation and the regression function simultaneously, potentially leading to better performance. For example, \cite{liu2020im2elevation} proposed an end-to-end CNN model for estimating the height of buildings from aerial images, while \cite{olenskyj2022end} used an end-to-end CNN for predicting crop yield from UAV-based multispectral images.

Although CNNs make it possible to carry out advanced classification or regression operations from input images, certain tasks require more complex models, such as Transformer models \cite{vaswani2017attention}.

Transformer models were first introduced by \cite{vaswani2017attention} in the context of machine translation and have been widely used in recent years for various NLP tasks due to their ability to model long-term dependencies and capture contextual relationships. One of the most popular applications of transformers is in language modeling, where the goal is to predict the next word in a sequence given the previous words. It has also been applied to a wide range of other NLP tasks, including text classification, sentiment analysis and question answering. \cite{devlin2018bert} and \cite{antol2015vqa} present transformers models to perform on these tasks.

With the images we seek to analyze, a CNN is sufficient to extract information from the image \cite{nibali2018numerical}. However, for our future work we will need models that are flexible and can potentially support multiple inputs. This is why we focused on advanced models capable of analyzing images and incorporating a transformer, a type of machine learning model well known for its flexibility and its ability to take different types of data as input(e.g, images and text). Composite models, also known  as hybrid models or end-to-end models are currently used in different fields such as medical field, to analyze medical images or in the field of drones to perform object recognition and classification from UAVs images. Recently google present \cite{ram2023multitask} using multi-task end-to-end transformer model on the task of conversational recommendations. which aim to provide recommendations based on a user’s explicit preferences expressed in dialogue.

Several studies have explored the integration of CNNs or transformer models for image classification and object detection tasks. For example, the Vision Transformer model proposed by \cite{kim2021vilt} uses a transformer-based architecture to process image patches and achieve state-of-the-art performance on image classification tasks. Similarly, the Detection Transformer proposed by \cite{carion2020end} uses a CNN to extract image features and predict object class in images. There is also the Hybrid CNN-Transformer model proposed by \cite{lu2023cnn} that combines a CNN and a transformer model for object detection tasks and achieves competitive results on benchmark datasets. 

The remainder of this article is organized as follows. Section 2 describes the compared models and Section 3 present the experiments. The results and our discussion are in in Section 4. Finally, Section 5 concludes this study.

\section{Proposed models}
This section introduces the models under comparison, namely the chained models, the composite model, and the standalone CNN as a baseline. We outline the specific characteristics of the input data and elaborate on the operational principles of each model.

\subsection{Chained models}
\begin{figure}
  \centering
  \includegraphics[width=0.75\textwidth]{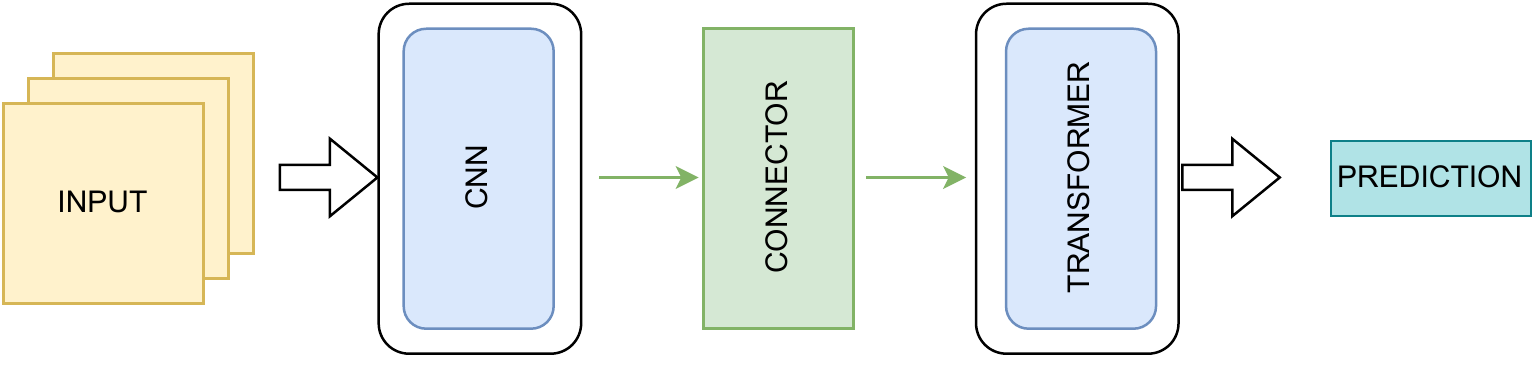}
  \caption{Chained models architecture }
  \label{fig:fig1}
\end{figure}

Chained models, as depicted in Figure \ref{fig:fig1}, are composed of two independently trained models, a CNN and a Transformer. Given that a transformer operates on tokenized inputs, we also add a connector between these models whose role is to sequence and tokenize the CNN outputs to feed into the Transformer. The tokenization is performed character by character with a vocabulary composed of 14 characters: [0,1,2,3,4,5,6,7,8,9, ,-,e,:].

First, each input image of size H x W x C (H, W, and C respectively stand for height, weight, and channel number) is input into the CNN to produce 2 x B + 2 x U output values (B and U respectively represent the number of base stations and the number of UAVs in the image). Each of these output values represents an 'x' or 'y' coordinate of the different elements present in each image (the squares and the triangle). They are tokenized to be put as input to the transformer.

The transformer model is trained with the (x, y) coordinates of the squares and the triangle as inputs and with the (x, y) coordinates of the square closest to the triangle as the label. The goal is to predict the final (x, y) value representing the coordinates of the target base station. By training the CNN with the images as input and the (x, y) coordinates of the squares and the triangle as labels, and subsequently training the transformer with the (x, y) coordinates of the squares and the triangle as inputs and the position of the closest square as the label, the model learns to generate accurate predictions of the target base station's coordinates.

\subsection{Composite model}
\begin{figure}
  \centering
  \includegraphics[width=0.75\textwidth]{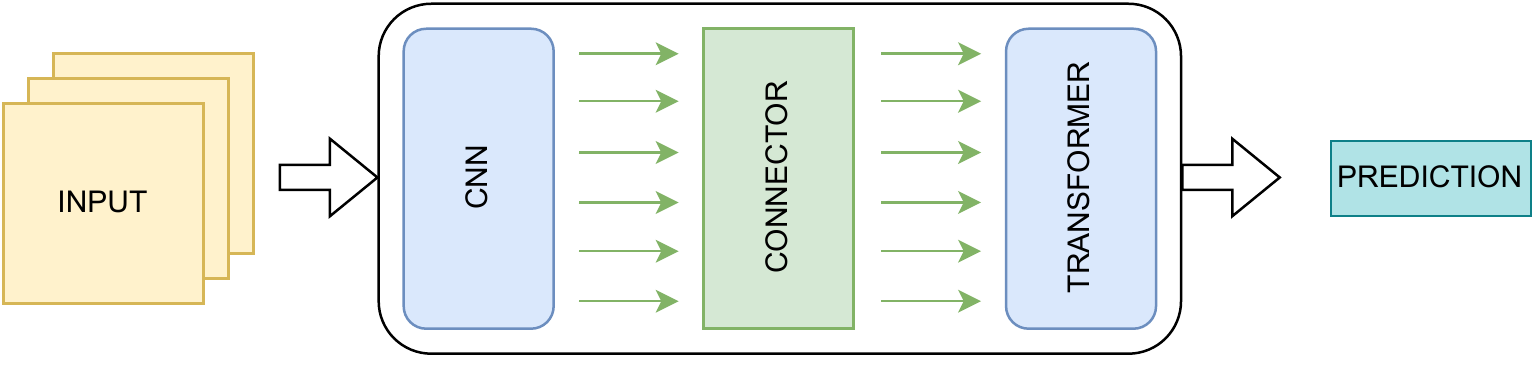}
  \caption{Composite model architecture }
  \label{fig:fig2}
\end{figure}

The composite model, as depicted in Figure \ref{fig:fig2}, is quite similar to the chained one but is trained end-to-end. 

In this composite model, we use input images of size H x W x C. The CNN and the Transformer are components of the same model and are trained together. Unlike the previous model, the number of CNN outputs can be customized as a hyper-parameter, providing flexibility. This integration of the CNN and Transformer creates a black box-like model, where specific parameters can be modified to enhance performance.

During training, the model is fed with samples comprising an image as the input feature and the (x, y) coordinates of the square closest to the triangle as the label. The objective is to train the model to accurately predict the coordinates of the target base station. By combining the capabilities of the CNN and the Transformer in a single model and optimizing the parameters, we aim to improve the model's performance.

\subsection{CNN}
\begin{figure}
  \centering
  \includegraphics[width=0.75\textwidth]{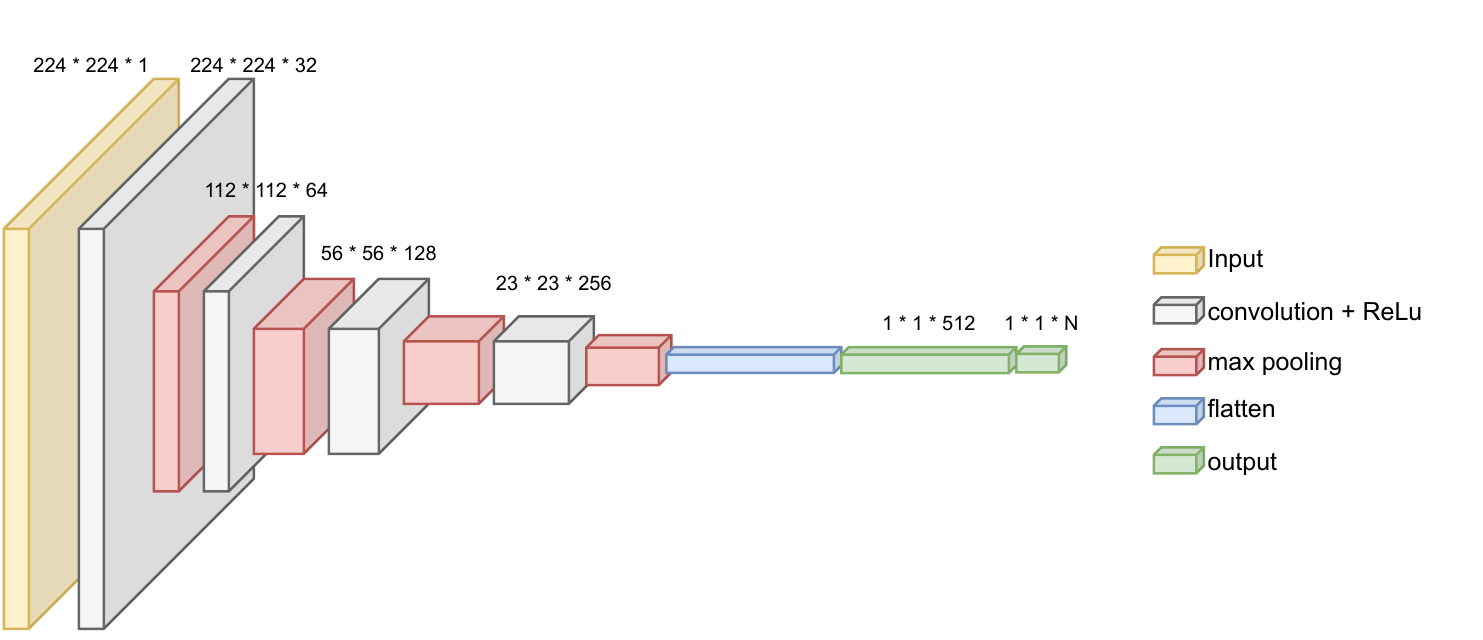}
  \caption{detail of the CNN of our models}
  \label{fig:fig3}
\end{figure}

We used a simplified version of VGG16 \cite{qassim2018compressed} as shown in Figure \ref{fig:fig3}. Our CNN is composed of four convolutional layers with maximum pooling. The input images are of size H x W x C, and the number of CNN outputs N, is set to 2 to compare our result to the ground truth as a coordinate point.

\section{Experiment}
\label{sec:headings}
This section presents how we optimized our models and how we evaluated them. We specify the hyper-parameters that have been the most influential as well as their optimal values for our use case. We also detail the dataset we use and what composes it.

\subsection{Dataset}
As we have chosen a simplify task, the dataset is also simplified. The dataset we used is entirely synthetic and specific to this problem. This allowed us to create a large amount of data that can be used to train both a CNN and a transformer. The dataset is composed of 1,000,000 images of size 224*224 in gray-scale and the labels associated with it. Some examples can be found in the appendix. The labels associated with each image contain location features that can be used to train our models. Each sample contains a relative path to the relevant image, the (x, y) coordinates of a first square, the (x, y) coordinates of a second square, the (x, y) coordinates of a triangle, and finally, as the ground-truth (x, y) coordinates of the nearest square to the triangle. 

Classically for each training, we separated the dataset into two parts: we kept 1000 images for evaluation and used to the other images to train our models. We also created an additional dataset of 1000 images to test our models.

\subsection{Metrics}
To evaluate the efficiency of our models, we measured the root mean square error (RMSE) between the point predicted by our model and the ground-truth available in the test dataset. We measured both the average RMSE and the standard deviation to obtain an accuracy metric. In our experiment, the RMSE is between 0 and $\sqrt{2}$, a lower average RMSE indicates a better fit between the model and the data.

\subsection{Hyper-parameters}
To fairly compare the models, we used the same hardware, we used a single Nvidia Geforce RTX 2080 TI. And we chose to tune hyper-parameters that are well known in machine learning to be important, as well as hyper-parameters that have a direct influence on our model.

We modified the learning rate of our model and the size of the dataset to train our models. Two other important hyper-parameters are the number of CNN outputs and the number of decimal places of these output values. They represent the verbosity of the CNN, the amount of information the CNN transfers to the transformer. And the last parameter we changed was the connector function whose role is to organize the CNN information before tokenization, which is essential to make the CNN output understandable by the transformer.

To optimize our models, we employed the Ray Tune library to simultaneously test various parameter combinations. The MLflow library played a crucial role in providing insights into the model's performance, allowing us to assess the impact of different parameter settings and facilitate comparisons between models. This approach allowed us to make informed decisions in selecting the most effective hyper-parameters to maximize the performance of our models.

For the chained models hyper-parameters, the number of values and decimal places in the CNN output were determined by the dataset used to construct the CNN. To compare the CNN predictions with the dataset ground truth values, we set the CNN output to 6 values with 3 decimal places. we only experiment with a connector function who converting into characters the CNN output float but adding a spaces character between values and a separator character between the context and the target during training.

To optimize the first model, we adjusted the learning rate and dataset size. A dataset size of 200,000 samples for the CNN and 1,000,000 samples for the Transformer struck a balance between training time and performance. Additionally, we determined that a learning rate of 1e-5 was most efficient for the CNN, while a learning rate of 3e-4 yielded optimal results for the Transformer.

For the composite model hyper-parameters, we experimented two methods to connect the CNN outputs into a sequence of tokens: the first one was the same we use for chained models and another prioritizing important information such as the initial digits of each CNN float value. However, the latter approach introduced more issues than solutions, so we kept the former.

We also optimized the learning rate of the composite model and found that a value around 1e-4 yielded good results. When it came to the number of decimal places in the CNN outputs, experimenting with 3, 6, and 8 decimals revealed that 6 decimals provided the best performance. Three decimals did not provide sufficient relevant information to the Transformer, while 8 decimals overwhelmed it. Six decimal places struck the right balance.

Regarding the number of values provided by the CNN, increasing this count improved performance, and we settled on 76 values as the output. However, further increases in this value resulted in longer training times and higher resource requirements.

Finally, we explored the impact of dataset size on model optimization. Increasing the dataset size led to better results, but we were limited by time and available resources. Nevertheless, we were able to conduct experiments with 1,000,000 samples.

\section{Results and discussion}

\begin{table}[h!]
  \normalsize
  \centering
  \begin{tabular}{|c|c|c|c|}
    \hline
    Models & RMSE & Training time & Training labels \\
    \hline
    Chained models & $0.055 \pm 0.120$  & 4 hours &  T*N \\ 
    \hline
    Composite model & $0.056 \pm 0.113$ & 2 days & N\\
    \hline
    CNN (baseline) & $0.061 \pm 0.055$ & 2.6 hours & N\\
    \hline
  \end{tabular}
  \caption{Results}
  \label{tab:cooperation_metrics_results}
\end{table}

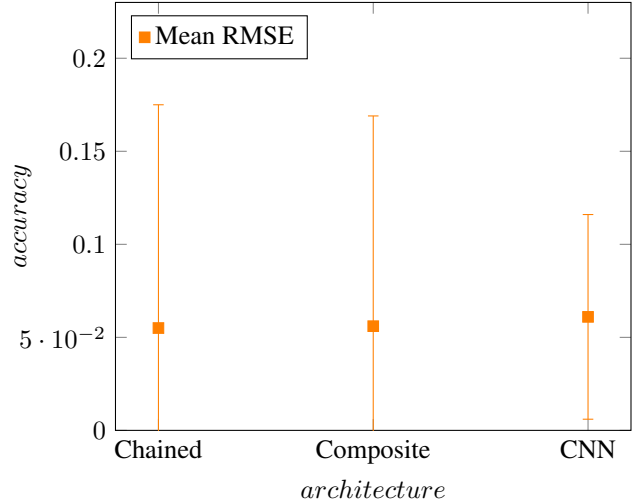
\begin{figure}
  \centering
  \begin{tikzpicture}
    \begin{axis}[
      xlabel={$architecture$},
      ylabel={$accuracy$},
      legend pos=north west,
      ymin=0,
      ymax=0.23,
      symbolic x coords={Chained, Composite, CNN},
      xtick={Chained, Composite, CNN},
      legend image post style={mark=square*, draw=orange, fill=orange}
      ]

      \addplot+[only marks, mark=square*, mark options={color=orange, fill=orange}, error bars/.cd, y dir=both, y explicit, error bar style={orange}]
      coordinates {
        (Chained, 0.055) +- (0, 0.12)
        (Composite, 0.056) +- (0, 0.113)
        (CNN, 0.061) +- (0, 0.055)
      };

      \legend{Mean RMSE}
    \end{axis}
  \end{tikzpicture}
  \label{fig:fig4}
  \caption{RMSE of the tested models}
\end{figure}

In this paper we compared 3 models: the chained models, the composite model and the CNN.

The chained models implied to split the task into sub-tasks and required additional labeling effort to train each model. The training lasted 2 hours for each model to accurately predict the position of the square closest to the triangle with an accuracy of $0.055 \pm $0.120.

The composite model allowed to avoid a large part of the labeling work by delegating the sub-task into a single so-called composite model. The training lasted 2 days and accurately predicted the position of the square closest to the triangle with an accuracy of $0.056 \pm $0.113. 

As a baseline, we also trained the CNN to extract only 2 values from the image, being the x y coordinates of the square closest to the triangle. We achieved an accuracy of $0.061 \pm $0.055 and a learning time of 2.6 hours. 

Compared to the baseline, the two models studied are equivalent  regarding the average RMSE values. However, they are slightly less efficient regarding the standard deviation. Keeping in mind our goal to address more complex use cases in the future (e.g, multi-modal models with text and images), both transformer-based models represent a suitable solution.

All these results are presented in the Table \ref{tab:cooperation_metrics_results} and Figure \ref{fig:fig4} and we also quantify the amount of labeling work needed for each model, here T represents the number of sub-tasks that must be performed by our models and N, the amount of work to label a sub-task. The quantification of labeling work holds significant importance for us, particularly because our future work will involve larger datasets that are more complex and challenging to label compared to the ones discussed in this article. Therefore, our objective is to minimize this labeling work to optimize our time and enhance overall efficiency. 

The results obtained from both the chained models and the composite model demonstrate that it is indeed possible to avoid the labeling work associated with training multitasking chained models, albeit at a notable additional cost. This finding is crucial for our future work as it enables us to concentrate more on fine-tuning the model itself, rather than solely focusing on the training datasets. By leveraging this information, we can allocate more resources towards optimizing and tuning the model for improved performance.

\section{Conclusion}
In this article, we compared chained models and composite model architectures. We observed that our chained models require more labeling effort than the composite model. Indeed, the composite model significantly reduces the labeling time by dividing it by the number of sub-tasks we aim to accomplish. Nevertheless, the composite model requires twelve times more training time to match the efficiency of the chained models. 

Both architectures offer fair performances regarding our use case with only $0.055 \pm $0.120 and $0.056 \pm $0.113 RMSEs between the predicted and ground truth values. When sub-tasks can be explicitly defined and when training labels for each sub-task are available, the chained models architecture is preferred as it requires less hyper-parameters, saves training time and therefore power energy. Otherwise, the composite model architecture is more suitable when sub-tasks lack clear delineation and can't be clearly defined; additionally, it also offers the perspective to address future use cases requiring multi-modal inputs and various tasks. 

Although further work is needed on other use cases to confirm our conclusion, these results prove essential for our future work, integrating multi-modal inputs and more elaborate images. 

\bibliographystyle{unsrt}  
\bibliography{bibPaper}

\section*{Appendix: distance calculation with GPT4.0}

This section presents a dialogue exchange (prompts and responses) based on multimodal inputs (text and images) performed with the GPT-4.0 model (as of April 14, 2023). For better readability, the URL prefix of PNG images has been removed from the dialogue.

\begin{itemize}
\item \emph{What does the image Full\_GPT\_architecture.png represent?}\newline
The image represents the architecture of a Full GPT (Generative Pre-trained Transformer) model. GPT is a type of transformer-based language model used for various natural language processing tasks, such as text generation, translation, and summarization. The image shows the different layers and components of the GPT model, including the input and output layers, self-attention mechanism, position-wise feed-forward networks, and layer normalization.

\item \emph{In image map\_0.png the point with coordinates x=0.000 and y=0.000 is at the bottom left of the image and the point with coordinates x=1.000 and y=1.000 is at the top right of the image. The triangle is positioned at x=0.622 and y=0.439; a first square is positioned at x=0.517 and y=0.898 and a second square is positioned at x=0.378 and y=0.886. The closest square to the triangle is positioned at x=0.517 and y=0.898. Is that clear for you?}\newline
Yes, that is clear. The triangle is at coordinates (0.622, 0.439), and there are two squares - one at (0.517, 0.898) and the other at (0.378, 0.886). The closest square to the triangle is the one at (0.517, 0.898).

\item \emph{In image map\_1.png the point with coordinates x=0.000 and y=0.000 is at the bottom left of the image and the point with coordinates x=1.000 and y=1.000 is at the top right of the image. The triangle is positioned at x=0.577 and y=0.713; a first square is positioned at x=0.778 and y=0.263 and a second square is positioned at x=0.583 and y=0.137. The closest square to the triangle is positioned at x=0.778 and y=0.263. Is that clear for you?}\newline
Yes, that is clear. In the image, the triangle is at coordinates (0.577, 0.713), and there are two squares - one at (0.778, 0.263) and the other at (0.583, 0.137). The closest square to the triangle is the one at (0.778, 0.263).

\item \emph{In image map\_2.png the point with coordinates x=0.000 and y=0.000 is at the bottom left of the image and the point with coordinates x=1.000 and y=1.000 is at the top right of the image. The triangle is positioned at x=0.127 and y=0.626; a first square is positioned at x=0.196 and y=0.341 and a second square is positioned at x=0.152 and y=0.243. The closest square to the triangle is positioned at x=0.196 and y=0.341. Is that clear for you?}\newline
Yes, that is clear. In the image, the triangle is at coordinates (0.127, 0.626), and there are two squares - one at (0.196, 0.341) and the other at (0.152, 0.243). The closest square to the triangle is the one at (0.196, 0.341).

\item \emph{In image map\_3.png the point with coordinates x=0.000 and y=0.000 is at the bottom left of the image and the point with coordinates x=1.000 and y=1.000 is at the top right of the image.  What is the position (x, y) of the square closest to the triangle?}\newline
In the image map\_3.png, the triangle is positioned at (0.372, 0.497). There are two squares - one at (0.430, 0.542) and the other at (0.283, 0.464). The closest square to the triangle is the one at (0.283, 0.464).

\item \emph{No, in the image map\_3.png the triangle is positioned at x=0.597 and y=0.730; a first square is positioned at x=0.693 and y=0.873 and a second square is positioned at x=0.323 and y=0.142. The closest square to the triangle is positioned at x=0.693 and y=0.873.}\newline
Apologies for the confusion. In the image map\_3.png, the triangle is positioned at (0.597, 0.730). There are two squares - one at (0.693, 0.873) and the other at (0.323, 0.142). The closest square to the triangle is the one at (0.693, 0.873).

\item \emph{In image map\_4.png the point with coordinates x=0.000 and y=0.000 is at the bottom left of the image and the point with coordinates x=1.000 and y=1.000 is at the top right of the image. What is the position (x, y) of the square closest to the triangle?}\newline
In the image map\_4.png, the triangle is positioned at (0.497, 0.372). There are two squares - one at (0.542, 0.430) and the other at (0.464, 0.283). The closest square to the triangle is the one at (0.464, 0.283).

\item \emph{Yes, you had almost the right answer. In image map\_4.png, the triangle is positioned at x=0.138 and y=0.135; a first square is positioned at x=0.564 and y=0.311 and a second square is positioned at x=0.597 and y=0.211. The closest square to the triangle is positioned at x=0.564 and y=0.311.}\newline
Apologies for the confusion. In the image map\_4.png, the triangle is positioned at (0.138, 0.135). There are two squares - one at (0.564, 0.311) and the other at (0.597, 0.211). The closest square to the triangle is the one at (0.564, 0.311).

\item \emph{In image map\_5.png the point with coordinates x=0.000 and y=0.000 is at the bottom left of the image and the point with coordinates x=1.000 and y=1.000 is at the top right of the image. What is the position (x, y) of the square closest to the triangle} \newline
In the image map\_5.png, the triangle is positioned at (0.497, 0.372). There are two squares - one at (0.542, 0.430) and the other at (0.464, 0.283). The closest square to the triangle is the one at (0.464, 0.283).

\item \emph{No, in image map\_5.png the square closest to the triangle is positioned at x=0.604 and y=0.148. }\newline
Apologies for the confusion. In the image map\_5.png, the triangle is positioned at (0.604, 0.148). The closest square to the triangle is also positioned at (0.604, 0.148).

\end{itemize}

To replay the prompts, use the following URL prefix: "https://raw.githubusercontent.com/marxav/aloa/main/". 

\begin{table}
\centering
\begin{tabular}{|c|c|c|c|c|c|}
\hline
Image & image\_name & triangle & squares[0] & squares[1] & target \\
\hline
\includegraphics[scale=0.25]{Figure/images/map\_0.png}
 & map\_0.png & (0.622, 0.439) & (0.517, 0.898) & (0.378, 0.886) & (0.517, 0.898) \\
\hline
\includegraphics[scale=0.25]{Figure/images/map\_1.png}
 & map\_1.png & (0.577, 0.713) & (0.778, 0.263) & (0.583, 0.137) & (0.778, 0.263) \\
\hline
\includegraphics[scale=0.25]{Figure/images/map\_2.png}
 & map\_2.png & (0.127, 0.626) & (0.196, 0.341) & (0.152, 0.243) & (0.196, 0.341) \\
\hline
\includegraphics[scale=0.25]{Figure/images/map\_3.png}
 & map\_3.png & (0.693, 0.873) & (0.323, 0.142) & (0.597, 0.730) & (0.693, 0.873) \\
\hline
\includegraphics[scale=0.25]{Figure/images/map\_4.png}
 & map\_4.png & (0.564, 0.311) & (0.597, 0.211) & (0.138, 0.135) & (0.564, 0.311) \\
\hline
\includegraphics[scale=0.25]{Figure/images/map\_5.png}
 & map\_5.png & (0.103, 0.382) & (0.604, 0.148) & (0.743 0.673) & (0.604, 0.148) \\
\hline
\end{tabular}
\caption{Images used to experiment with GPT4.0}
\label{tab:tab-GPT}
\end{table}
\end{document}